\documentclass[11pt,letterpaper]{article}
\usepackage[letterpaper]{geometry}
\usepackage{acl2012}
\usepackage{times}
\usepackage{latexsym}
\usepackage{amsmath}
\usepackage{multirow}
\usepackage{url}
\usepackage[table]{xcolor}
\usepackage{underscore}
\usepackage{tikz}
\usepackage{ifthen}
\usepackage{pgfplots}
\usepackage{soul}

\makeatletter
\newcommand{\@BIBLABEL}{\@emptybiblabel}
\newcommand{\@emptybiblabel}[1]{}
\makeatother
\usepackage[hidelinks]{hyperref}

\pgfplotsset{compat=1.10}
\usetikzlibrary{bayesnet}
\definecolor{leaRed}{rgb}{0.722, 0, 0.278}
\definecolor{leaGreen}{rgb}{0.13, 0.67, 0.8}

\title{Whodunnit? Crime Drama as a Case for Natural Language Understanding}

\author{Lea Frermann \hspace{2ex} Shay B. Cohen \hspace{2ex} Mirella Lapata\\
  Institute for Language, Cognition and Computation\\
  School of Informatics, University of Edinburgh\\
  10 Crichton Street, Edinburgh EH8 9AB
\\
  {\tt l.frermann@ed.ac.uk \hspace{1ex} scohen@inf.ed.ac.uk  \hspace{1ex} mlap@inf.ed.ac.uk} }

\date{}

\begin{document}
\maketitle
\begin{abstract}
  In this paper we argue that crime drama exemplified in television
  programs such as \emph{CSI: Crime Scene Investigation} is an ideal
  testbed for approximating real-world natural language understanding
  and the complex inferences associated with it. We propose to treat
  crime drama as a new inference task, capitalizing on the fact that
  each episode poses the same basic question (i.e.,~who committed the
  crime) and naturally provides the answer when the perpetrator is
  revealed.  We develop a new dataset\footnote{Our dataset is
    available at \url{https://github.com/EdinburghNLP/csi-corpus}.}
  based on CSI episodes, formalize perpetrator identification as a
  sequence labeling problem, and develop an LSTM-based model which
  learns from multi-modal data. Experimental results show that an
  incremental inference strategy is key to making accurate guesses as
  well as learning from representations fusing textual, visual, and
  acoustic input.
\end{abstract}

\section{Introduction}

The success of neural networks in a variety of applications
\cite{Sutskever:2014,vinyals2015show} and the creation of large-scale
datasets have played a critical role in advancing machine
understanding of natural language on its own or together with other
modalities. The problem has assumed several guises in the literature
such as reading comprehension
\cite{richardson-burges-renshaw:2013:EMNLP,rajpurkar-EtAl:2016:EMNLP2016},
recognizing textual entailment
\cite{bowman-EtAl:2015:EMNLP,rocktaschel2016reasoning}, and notably
question answering based on text \cite{Hermann:ea:2015,Weston:2015},
images \cite{antol2015vqa}, or video \cite{Tapaswietal:2016}.

In order to make the problem tractable and amenable to computational
modeling, existing approaches study isolated aspects of natural
language understanding. For example, it is assumed that understanding
is an offline process, models are expected to digest large amounts of
data before being able to answer a question, or make inferences. They
are typically exposed to non-conversational texts or still images when
focusing on the visual modality, ignoring the fact that understanding
is situated in time and space and involves interactions between
speakers. In this work we relax some of these simplifications by
advocating a new task for natural language understanding which
is multi-modal, exhibits spoken conversation, and is incremental,
i.e.,~unfolds sequentially in time. 

Specifically, we argue that crime drama exemplified in television
programs such as \emph{CSI: Crime Scene Investigation} can be used to
approximate real-world natural language understanding and the complex
inferences associated with it.  CSI revolves around a team of forensic
investigators trained to solve criminal cases by scouring the crime
scene, collecting irrefutable evidence, and finding the missing pieces
that solve the mystery.  Each episode poses the same ``whodunnit''
question and naturally provides the answer when the perpetrator is
revealed.  Speculation about the identity of the perpetrator is an
integral part of watching CSI and an incremental process: viewers
revise their hypotheses based on new evidence gathered around the
suspect/s or on new inferences which they make as the episode evolves.

%

We formalize the task of identifying the perpetrator in a crime series
as a sequence labeling problem. Like humans watching an episode, we
assume the model is presented with a sequence of inputs comprising
information from different modalities such as text, video, or audio
(see Section~\ref{sec:problem-formulation} for details). The model
predicts for each input whether the perpetrator is mentioned or
not. Our formulation generalizes over episodes and crime series. It is
not specific to the identity and number of persons committing the
crime as well as the type of police drama under consideration.
Advantageously, it is \emph{incremental}, we can track model
predictions from the beginning of the episode and examine its
behavior, e.g.,~how often it changes its mind, whether it is
consistent in its predictions, and when the perpetrator is identified.

We develop a new dataset based on~39 CSI episodes which contains
goldstandard perpetrator mentions as well as viewers' guesses about
the perpetrator while each episode unfolds. The sequential nature of
the inference task lends itself naturally to recurrent network
modeling. We adopt a generic architecture which combines a
one-directional long-short term memory network \cite{Hochreiter:1997}
with a softmax output layer over binary labels indicating whether the
perpetrator is mentioned. Based on this architecture, we investigate
the following questions:

\begin{enumerate}

\item What type of knowledge is necessary for  performing the
  perpetrator inference task? Is the textual modality sufficient or do
  other modalities (i.e.,   visual and auditory input) also  play a
  role? 

\item What type of inference strategy is appropriate? In other words,
  does access to past information matter for making accurate
  inferences?

\item To what extent does model behavior simulate humans? Does
  performance improve over time and how much of an episode does the
  model need to process in order to make accurate guesses?

\end{enumerate}

Experimental results on our new dataset reveal that multi-modal
representations are essential for the task at hand boding well with
real-world natural language understanding. We also show that an
incremental inference strategy is key to guessing the perpetrator
accurately although the model tends to be less consistent compared to
humans.  In the remainder, we first discuss related work
(Section~\ref{sec:related-work}), then present our dataset
(Section~\ref{sec:csi}) and formalize the modeling problem
(Section~\ref{sec:problem-formulation}). We describe our experiments
in Section~\ref{sec:evaluation}.

\section{Related Work}
\label{sec:related-work}

\begin{figure*}[t]
\includegraphics[width=\textwidth]{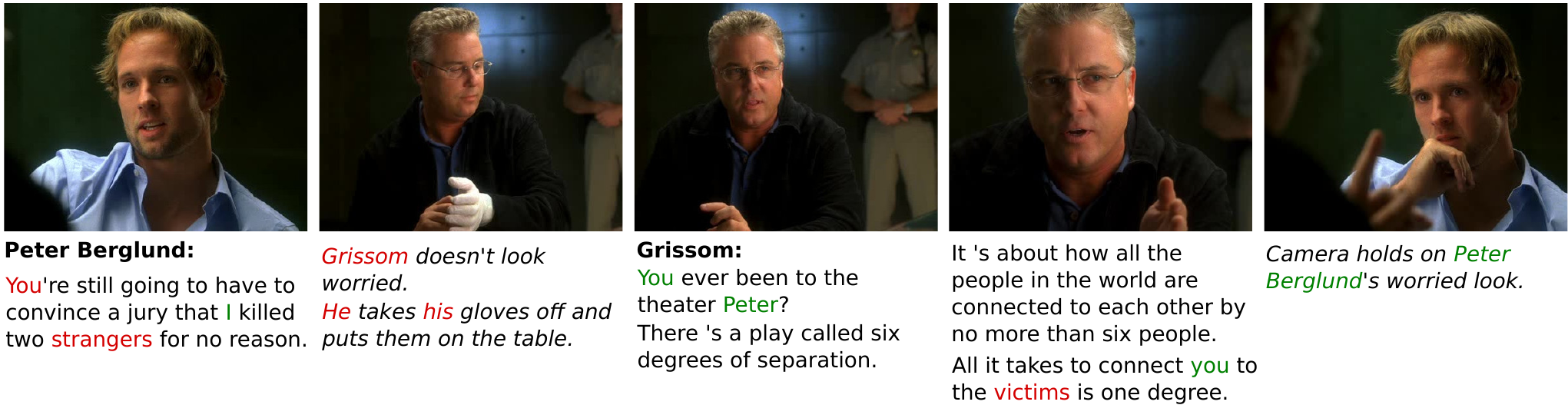}
\vspace*{-.5cm}
\caption{Excerpt from a CSI script (Episode~03, Season~03: ``Let the
  Seller Beware''). Speakers are shown in bold, spoken dialog in
  normal font, and scene descriptions in italics. Gold-standard entity
  mention annotations are in color. Perpetrator mentions (e.g.,~Peter
  Berglund) are in green, while words referring to other entities are
  in red.}
 \label{fig-script}
\end{figure*}

Our research has connections to several lines of work in natural
language processing, computer vision, and more generally multi-modal
learning. We review related literature in these areas below.

\paragraph{Language Grounding}

Recent years have seen increased interest in the problem of grounding
language in the physical world. Various semantic space models have
been proposed which learn the meaning of words based on linguistic and
visual or acoustic input
\cite{Bruni:2014,silberer:ea:16,lazaridou-pham-baroni:2015:NAACL-HLT,kiela-bottou:2014:EMNLP2014}.
A variety of cross-modal methods which fuse techniques from image and
text processing have also been applied to the tasks of generating
image descriptions and retrieving images given a natural language
query \cite{vinyals2015show,xu2015show,karpathy2015deep}. Another
strand of research focuses on how to explicitly encode the underlying
semantics of images making use of structural representations
\cite{ortiz2015learning,elliott2013image,yatskar2016situation,johnson2015image}.
Our work shares the common goal of grounding language in additional
modalities. Our model is, however, not static, it learns
representations which evolve over time.


\paragraph{Video Understanding} 

Work on video understanding has assumed several guises such as
generating descriptions for video clips
\cite{Venugopalan:2015,Venugopalan:2015naacl}, retrieving video clips
with natural language queries \cite{Lin:ea:2014}, learning actions in
video \cite{Bojanowski:ea:2013}, and tracking characters
\cite{Sivic:ea:2009}. Movies have also been aligned to screenplays
\cite{Cour:ea:2008}, plot synopses \cite{Tapaswi:ea:2015}, and books
\cite{Zhu:ea:2015} with the aim of improving scene prediction and
semantic browsing. Other work uses low-level features (e.g., based on
face detection) to establish social networks of main characters in
order to summarize movies or perform genre classification
\cite{Rasheed:2005,Sang:2010,Dimitrova:2000}. Although visual features
are used mostly in isolation, in some cases they are combined with
audio in order to perform video segmentation \cite{Boreczky:1998} or
semantic movie indexing \cite{Naphide:2001}.

A few datasets have been released recently which include movies and
textual data. MovieQA \cite{Tapaswietal:2016} is a large-scale dataset
which contains~408 movies and 14,944 questions, each accompanied with
five candidate answers, one of which is correct. For some movies, the
dataset also contains subtitles, video clips, scripts, plots, and text
from the Described Video Service (DVS), a narration service for the
visually impaired. MovieDescription \cite{Rohrbach:ea:2017} is a
related dataset which contains sentences aligned to video clips from
200 movies. Scriptbase \cite{Gorinski:2015} is another movie database
which consists of movie screenplays (without video) and has been used
to generate script summaries.

In contrast to the story comprehension tasks envisaged in MovieQA and
MovieDescription, we focus on a single cinematic genre (i.e.,~crime
series), and have access to \emph{entire} episodes (and their
corresponding screenplays) as opposed to video-clips or DVSs for some
of the data. Rather than answering multiple factoid questions, we aim to
solve a \emph{single} problem, albeit one that is inherently challenging
to both humans and machines.


\paragraph{Question Answering}

A variety of question answering tasks (and datasets) have risen in
popularity in recent years. Examples include reading comprehension,
i.e.,~reading text and answering questions about it
\cite{richardson-burges-renshaw:2013:EMNLP,rajpurkar-EtAl:2016:EMNLP2016},
open-domain question answering, i.e.,~finding the answer to a question
from a large collection of documents
\cite{Voorhees:Tice:2000,yang-yih-meek:2015:EMNLP}, and cloze question
completion, i.e.,~predicting a blanked-out word of a sentence
\cite{Hill:ea:2015,Hermann:ea:2015}. Visual question answering (VQA;
\newcite{antol2015vqa}) is a another related task where the aim is to
provide a natural language answer to a question about an image.

Our inference task can be viewed as a form of question answering over
multi-modal data, focusing on one type of question. Compared to
previous work on machine reading or visual question answering, we are
interested in the temporal characteristics of the inference process,
and study how understanding evolves incrementally with the
contribution of various modalities (text, audio, video). Importantly,
our formulation of the inference task as a sequence labeling problem
departs from conventional question answering allowing us to study how
humans and models alike make decisions over time.

\section{The CSI Dataset}
\label{sec:csi}

In this work, we make use of episodes of the U.S. TV show ``Crime
Scene Investigation Las Vegas'' (henceforth CSI), one of the most
successful crime series ever made. Fifteen seasons with a total of~337
episodes were produced over the course of fifteen years. CSI is a
procedural crime series, it follows a team of investigators employed
by the Las Vegas Police Department as they collect and evaluate evidence
to solve murders, combining forensic police work with the
investigation of suspects.


We paired official CSI videos (from seasons~\mbox{1--5}) with
screenplays which we downloaded from a website hosting TV show
transcripts.\footnote{\url{http://transcripts.foreverdreaming.org/}}
Our dataset comprises 39~CSI episodes, each approximately
43~minutes long. Episodes follow a regular plot, they begin with the
display of a crime (typically without revealing the perpetrator) or a
crime scene.  A team of five recurring police investigators attempt to
reconstruct the crime and find the perpetrator. During the
investigation, multiple (innocent) suspects emerge, while the crime is
often committed by a single person, who is eventually identified and
convicted. Some CSI episodes may feature two or more unrelated
cases. At the beginning of the episode the CSI team is split and each
investigator is assigned a single case. The episode then alternates
between scenes covering each case, and the stories typically do not
overlap.  Figure~\ref{fig-script} displays a small excerpt from a CSI
screenplay.  Readers unfamiliar with script writing conventions should
note that scripts typically consist of scenes, which have headings
indicating where the scene is shot (e.g., inside someone's
house). Character cues preface the lines the actors speak (see
boldface in Figure~\ref{fig-script}), and scene descriptions explain
what the camera sees (see second and fifth panel in
Figure~\ref{fig-script}).

\begin{table}[t]
\begin{small}
 \begin{tabular}{|lllrrr|}
 \hline
 & \multicolumn{2}{l}{episodes with one case} & 19&& \\
 &  \multicolumn{2}{l}{episodes with two cases}& 20&& \\
 &  \multicolumn{2}{l}{total number of cases}   & 59&& \\\hline\hline
 &                              & & min & max & avg\\
\multirow{4}{*}{\rotatebox{90}{per case}}  &  \multicolumn{2}{l}{sentences}  & 228 & 1209 & 689\\
 & \multicolumn{2}{l}{{sentences with perpetrator}} &0 &267 & 89 \\
 &  \multicolumn{2}{l}{scene descriptions} & 64&538&245 \\
 &  \multicolumn{2}{l}{spoken utterances} & 144&778&444\\
 &  \multicolumn{2}{l}{characters}          & 8& 38& 20 \\\hline\hline
 &  \multirow{4}{*}{type of crime}   & murder     &51&& \\
 &                  & accident   & 4&&\\
 &                  & suicide    & 2&&\\
 &                  & other      & 2&&\\\hline
 \end{tabular}
\end{small}
\caption{Statistics on the CSI data set.  The type of
  crime was identified by our annotators via a multiple-choice questionnaire
  (which included the option ``other''). Note that accidents may also
  involve perpetrators.} 
 \label{tab-data-stats}
\end{table}

Screenplays were further synchronized with the video using closed
captions which are time-stamped and provided {in the form of
  subtitles} as part of the video data. The alignment between
screenplay and closed captions is non-trivial, since the latter only
contain dialogue, omitting speaker information or scene
descriptions. We first used dynamic time warping (DTW;
\newcite{Myers:ea:1981}) to approximately align closed captions with
the dialogue in the scripts.  And then heuristically time-stamped
remaining elements of the screenplay (e.g.,~scene descriptions),
allocating them to time spans between spoken
utterances. Table~\ref{tab-data-stats} shows some descriptive
statistics on our dataset, featuring the number of cases per episode,
its length (in terms of number of sentences), the type of crime, among other information.

\begin{figure}[t]
\begin{footnotesize}
\hspace*{-.1cm}\begin{tabular}{|@{~}p{4cm}p{1.6cm}p{1cm}@{}|} \hline
\multicolumn{3}{|@{~}p{7.8cm}|}{\textbf{Number of cases: 2}} \\
\multicolumn{3}{|@{~}p{7.8cm}|}{Case 1: Grissom, Catherine, Nick and Warrick
  investigate when a wealthy couple is murdered at their house.} \\
\multicolumn{3}{|@{~}p{7.8cm}|}{Case 2: Meanwhile Sara is sent to a local high
  school where a cheerleader was found eviscerated on the football
  field.} \\ 
\multicolumn{3}{|c|}{} \\
\multicolumn{3}{|c|}{\includegraphics[scale=.33]{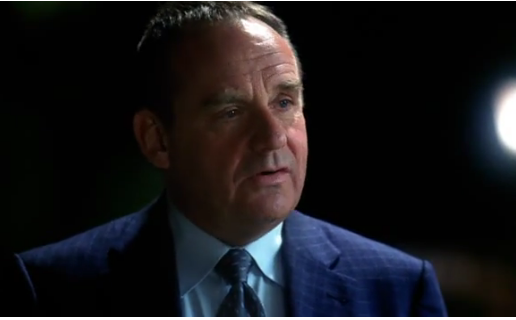}}\\
\multicolumn{3}{|c|}{\raisebox{1.9ex}[0pt]{\includegraphics[scale=.31]{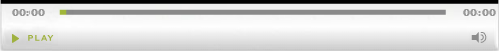}}\raisebox{1.9ex}[0pt]{\includegraphics[scale=.115]{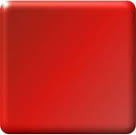}}} \\
\multicolumn{1}{|c}{\textbf{Screenplay}}& \textbf{Perpetrator mentioned?} & \textbf{Relates to
                                                case
                                                1/2/none?}\\ 
(Nick cuts the canopy around MONICA NEWMAN.) & \multicolumn{1}{c}{\raisebox{-2ex}[0pt]{\includegraphics[scale=.022]{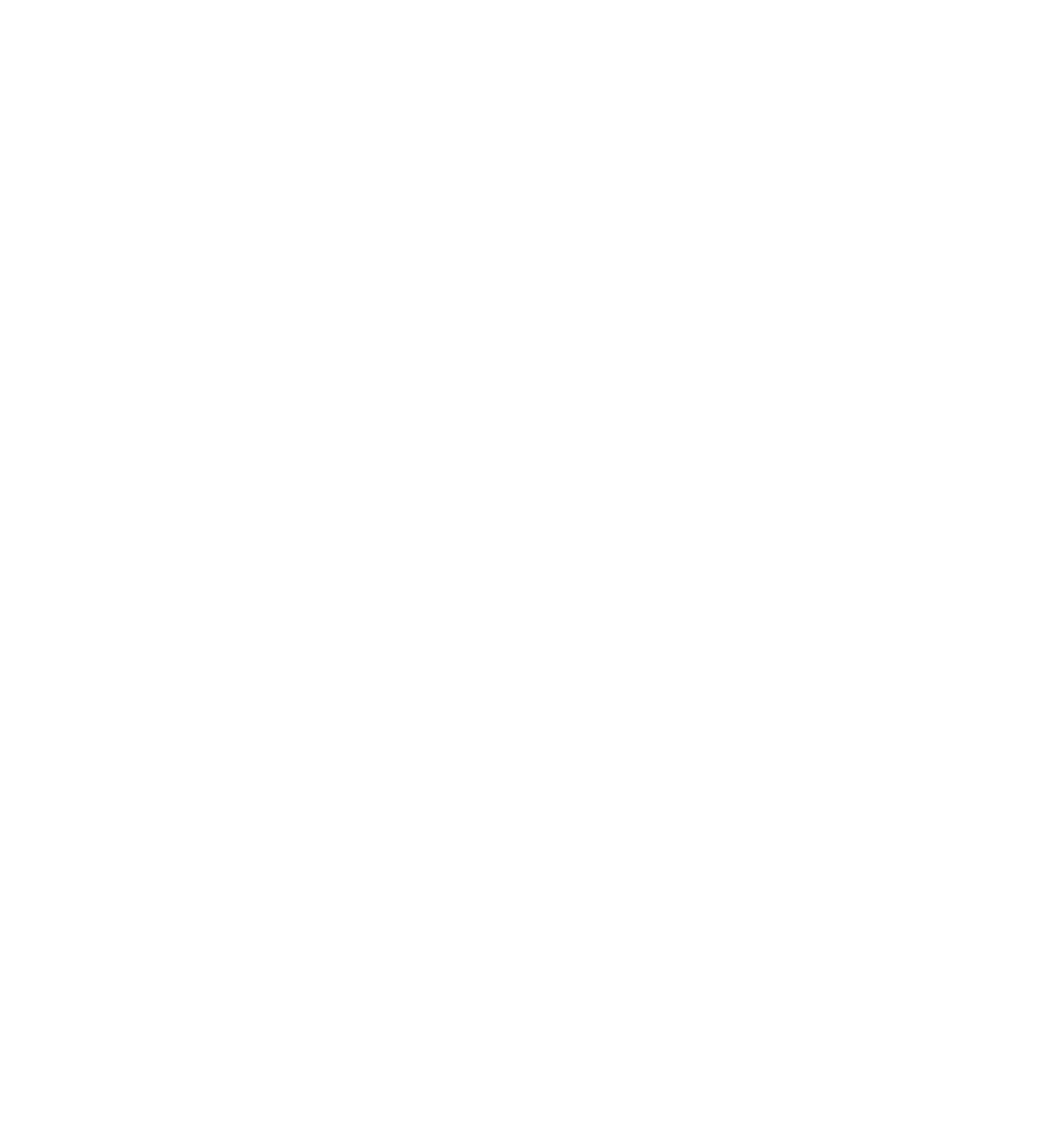}}}& \multicolumn{1}{c|}{\raisebox{-2ex}[0pt]{\includegraphics[scale=.045]{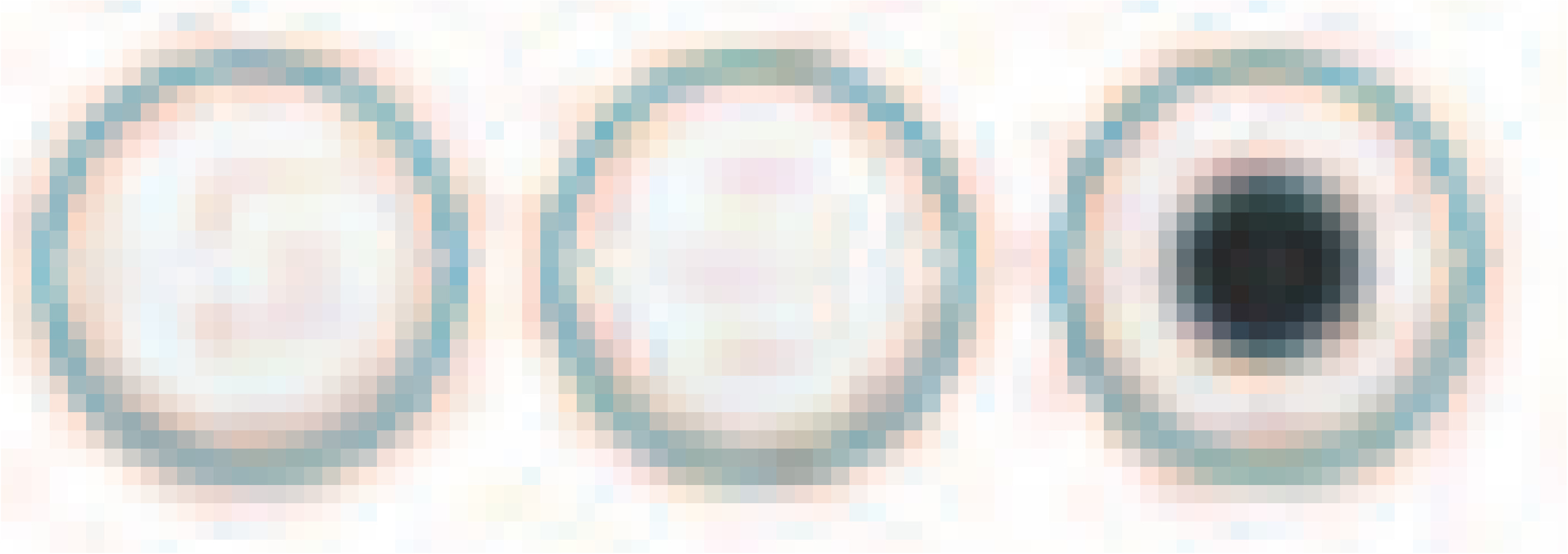}}}\\
\textbf{Nick} okay, Warrick, hit it &  \multicolumn{1}{c}{\raisebox{-2ex}[0pt]{\includegraphics[scale=.022]{pictures/box-crop.pdf}}}& \multicolumn{1}{c|}{\raisebox{-2ex}[0pt]{\includegraphics[scale=.045]{pictures/radio-crop.pdf}}}\\
(WARRICK starts the crane support under the awning to remove the body
  and the canopy area that NICK cut.) &
                                       \multicolumn{1}{c}{\raisebox{-2ex}[0pt]{\includegraphics[scale=.022]{pictures/box-crop.pdf}}}& \multicolumn{1}{c|}{\raisebox{-2ex}[0pt]{\includegraphics[scale=.045]{pictures/radio-crop.pdf}}}\\
\textbf{Nick} white female, multiple bruising $\dots$ bullet hole to
  the temple doesn't help &\multicolumn{1}{c}{\raisebox{-2ex}[0pt]{\includegraphics[scale=.022]{pictures/box-crop.pdf}}}& \multicolumn{1}{c|}{\raisebox{-2ex}[0pt]{\includegraphics[scale=.045]{pictures/radio-crop.pdf}}}\\
\textbf{Nick} .380 auto on the side &
                                      \multicolumn{1}{c}{\raisebox{-2ex}[0pt]{\includegraphics[scale=.022]{pictures/box-crop.pdf}}}&
                                                                                                                                     \multicolumn{1}{c|}{\raisebox{-2ex}[0pt]{\includegraphics[scale=.045]{pictures/radio-crop.pdf}}}\\
\textbf{Warrick} yeah, somebody man-handled her pretty good before
  they killed her &
                    \multicolumn{1}{c}{\raisebox{-2ex}[0pt]{\includegraphics[scale=.022]{pictures/box-crop.pdf}}}& \multicolumn{1}{c|}{\raisebox{-2ex}[0pt]{\includegraphics[scale=.045]{pictures/radio-crop.pdf}}}\\\hline
\end{tabular}
\end{footnotesize}
\vspace*{-.2cm}
\caption{\label{interface:a} Annotation interface (first pass): after
  watching three minutes of the episode,  the annotator indicates
  whether she believes the perpetrator has been mentioned.} 
\end{figure}

The data was further annotated, with two goals in mind. Firstly, in
order to capture the characteristics of the {\it human inference}
process, we recorded how participants incrementally update their
beliefs about the perpetrator. Secondly, we collected gold-standard
labels indicating whether the perpetrator is mentioned. Specifically,
while a participant watches an episode, we record their guesses about
who the perpetrator is (Section~\ref{ssec-annotation-behavior}). Once
the episode is finished and the perpetrator is revealed, the same
participant annotates entities in the screenplay referring to the {\it
  true} perpetrator~(Section~\ref{ssec-annotation-gold}).




\subsection{Eliciting Behavioral Data}
\label{ssec-annotation-behavior}

All annotations were collected through a web-interface. We recruited
three annotators, all postgraduate students and proficient in English,
none of them regular CSI viewers. We obtained annotations for 39 episodes
(comprising 59 cases).

A snapshot of the annotation interface is presented in
Figure~\ref{interface:a}.  The top of the interface provides a short
description of the episode, i.e.,~ in the form of a one-sentence
summary (carefully designed to not give away any clues about the
perpetrator). Summaries were adapted from the CSI season summaries
available in Wikipedia.\footnote{See e.g.,
  \url{https://en.wikipedia.org/wiki/CSI:_Crime_Scene_Investigation_(season_1)}.}
The annotator watches the episode (i.e.,~the video without closed
captions) as a sequence of three minute intervals. Every three
minutes, the video halts, and the annotator is presented with the
screenplay corresponding to the part of the episode they have just
watched. While reading through the screenplay, they must indicate for
every sentence whether they believe the perpetrator is mentioned. This
way, we are able to monitor how humans create and discard hypotheses
about perpetrators \emph{incrementally}. As mentioned earlier, some
episodes may feature more than one case. Annotators signal for each
sentence, which case it belongs to or whether it is irrelevant (see the
radio buttons in Figure~\ref{interface:a}). In order to obtain a more
fine-grained picture of the human guesses, annotators are additionally
asked to press a large red button (below the video screen) as soon as
they ``think they know who the perpetrator is'', i.e.,~at any time while they are watching the video. They are allowed to press
the button multiple times throughout the episode in case they change
their mind.


%
%

Even though the annotation task just described reflects individual
rather than gold-standard behavior, we report inter-annotator
agreement (IAA) as a means of estimating variance amongst
participants. We computed IAA using Cohen's \shortcite{Cohen:1960}
Kappa based on three episodes annotated by two participants. Overall
agreement on this task (second column in
Figure~\ref{interface:a}) is~0.74. We also measured percent agreement
on the minority class (i.e., sentences tagged as ``perpetrator
mentioned'') and found it to be reasonably good at 0.62, indicating
that despite individual differences, the process of guessing the
perpetrator is broadly comparable across participants.  Finally,
annotators had no trouble distinguishing which utterances refer to
which case (when the episode revolves around several), achieving an
IAA of~$\kappa=0.96$.



\begin{figure}[t]
\begin{footnotesize}

\begin{tabular}{|@{}l@{\hspace*{-.2cm}}c@{~}c@{~}c@{~}c@{~}c@{~}c@{~}c@{~}c@{}|} \hline
& ( & It & 's & a & shell & casing & . &) \\
\raisebox{5ex}[0pt]{Perpetrator} & \includegraphics[scale=0.035]{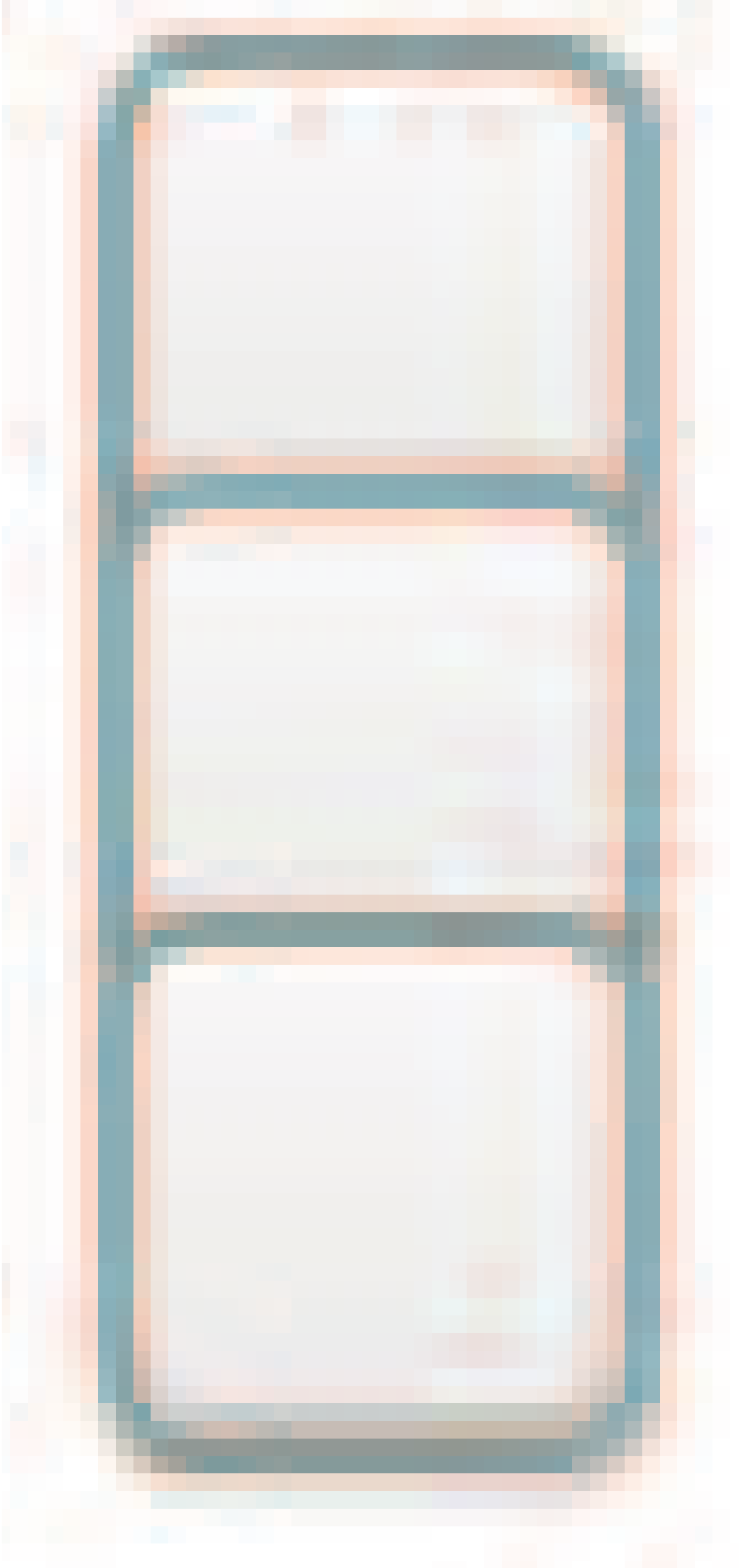}& \includegraphics[scale=0.035]{pictures/no-tick-crop.pdf} & \includegraphics[scale=0.035]{pictures/no-tick-crop.pdf}& \includegraphics[scale=0.035]{pictures/no-tick-crop.pdf}& \includegraphics[scale=0.035]{pictures/no-tick-crop.pdf}& \includegraphics[scale=0.035]{pictures/no-tick-crop.pdf}& \includegraphics[scale=0.035]{pictures/no-tick-crop.pdf}& \includegraphics[scale=0.035]{pictures/no-tick-crop.pdf}\\
\multicolumn{9}{|l|}{\raisebox{5.5ex}[0pt]{Suspect}} \\

\multicolumn{1}{|l}{\raisebox{6ex}[0pt]{Other}}&  GRISSOM & moves & his & light & to & the & canopy & below \\
\raisebox{5ex}[0pt]{Perpetrator} & \includegraphics[scale=0.0355]{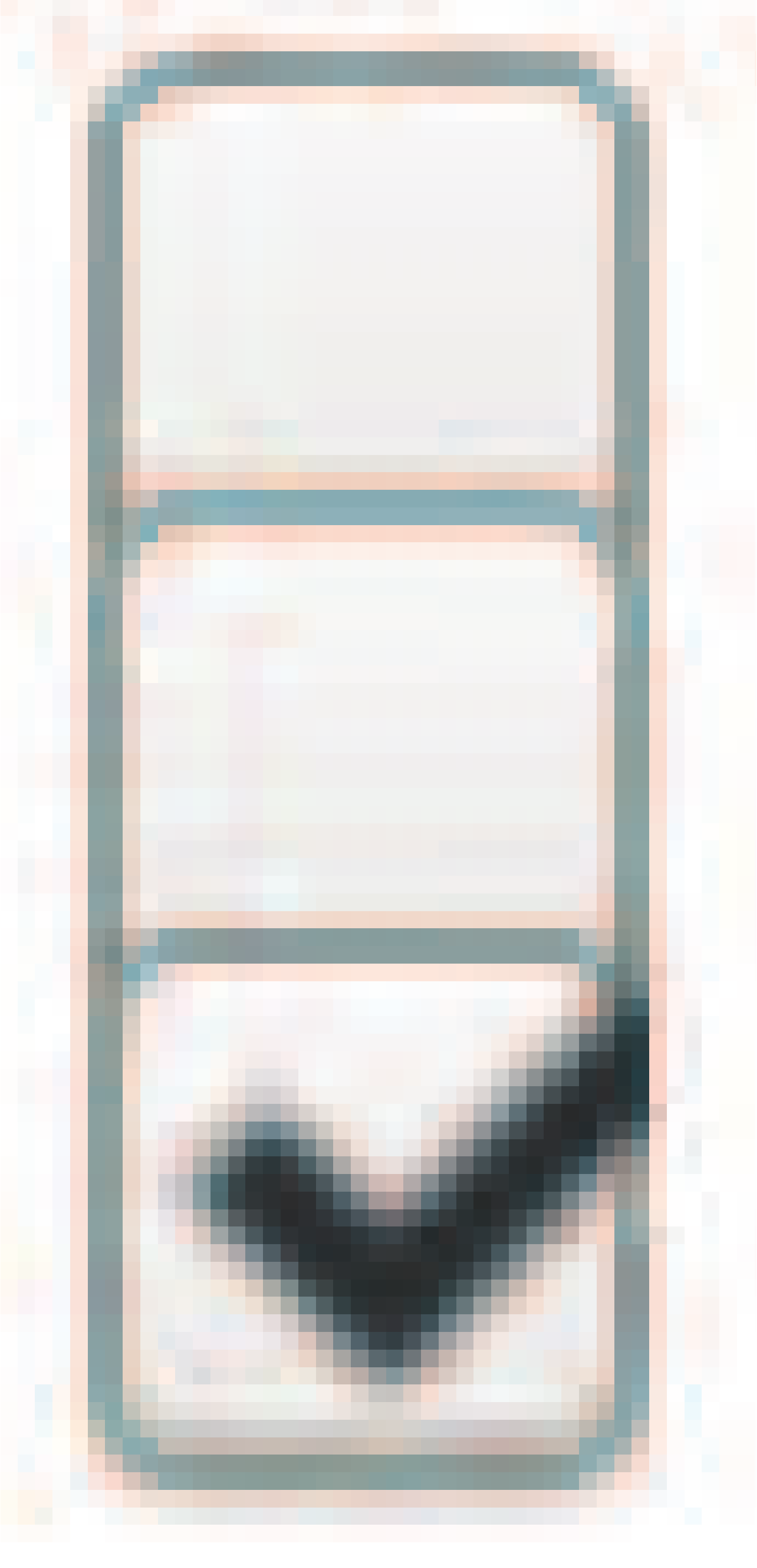}& \includegraphics[scale=0.035]{pictures/no-tick-crop.pdf} & \includegraphics[scale=0.035]{pictures/no-tick-crop.pdf}& \includegraphics[scale=0.035]{pictures/no-tick-crop.pdf}& \includegraphics[scale=0.035]{pictures/no-tick-crop.pdf}& \includegraphics[scale=0.035]{pictures/no-tick-crop.pdf}& \includegraphics[scale=0.035]{pictures/no-tick-crop.pdf}& \includegraphics[scale=0.035]{pictures/no-tick-crop.pdf}\\
\multicolumn{9}{l}{\raisebox{5.5ex}[0pt]{Suspect}} \\
\multicolumn{9}{l}{\raisebox{6ex}[0pt]{Other}}  \vspace*{-.8cm}\\\hline
\end{tabular}
\end{footnotesize}
\vspace*{-.2cm}
\caption{\label{interface:b} Annotation interface (second pass): after
  watching the episode, the annotator indicates for each word whether
  it refers to the perpetrator. }
\end{figure}

\subsection{Gold Standard Mention Annotation}
\label{ssec-annotation-gold}

After watching the entire episode, the annotator reads through the
screenplay for a second time, and tags entity mentions, now knowing
the perpetrator. Each word in the script has three radio buttons
attached to it, and the annotator selects one only if a word refers to
a perpetrator, a suspect, or a character who falls into neither of these classes (e.g.,~a police
investigator or a victim). For the majority of words, no button will
be selected.  A snapshot of our interface for this second layer of
annotations is shown in Figure~\ref{interface:b}. To ensure
consistency, annotators were given detailed guidelines about what
constitutes an entity. Examples include proper names and their titles
(e.g.,~\textsl{Mr Collins, Sgt. O' Reilly}), pronouns
(e.g.,~\textsl{he, we}), and other referring expressions including
nominal mentions (e.g.,~\textsl{let's arrest the guy with the black
  hat}). 

Inter-annotator agreement based on three episodes and two annotators
was~$\kappa=0.90$ on the perpetrator class and $\kappa=0.89$ on other
entity annotations (grouping together suspects with other
entities). Percent agreement was~0.824 for perpetrators and 0.823 for
other entities. The high agreement indicates that the task is
well-defined and the elicited annotations reliable.  After the second
pass, various entities in the script are disambiguated in terms of
whether they refer to the perpetrator or other individuals.

Note that in this work we do not use the token-level gold standard
annotations directly. Our model is trained on sentence-level
annotations which we obtain from token-level ones, under the
assumption that a sentence mentions the perpetrator if it contains a
token that does.





\section{Model Description}
\label{sec:problem-formulation}

We formalize the problem of identifying the perpetrator in a crime
series episode as a sequence labeling task. Like humans watching an
episode, our model is presented with a sequence of (possibly
multi-modal) inputs, each corresponding to a sentence in the script,
and assigns a label~$l$ indicating whether the perpetrator is
mentioned in the sentence ($l=1$) or not ($l=0$).  The model is fully
incremental, each labeling decision is based solely on information
derived from \emph{previously} seen inputs.

%

We could have formalized our inference task as a multi-label
classification problem where labels correspond to characters in the
script. Although perhaps more intuitive, the multi-class framework
results in an output label space different for each episode which
renders comparison of model performance across episodes
problematic. In contrast, our formulation has the advantage of being
directly applicable to any episode or indeed any crime series.

\begin{figure}[t]
 \includegraphics[width=0.48\textwidth]{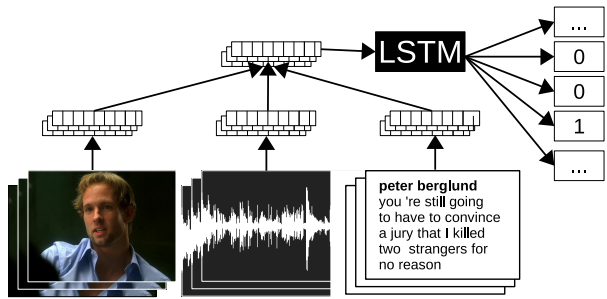}
\vspace*{-.2cm}
 \caption{Overview of the perpetrator prediction task. The model
   receives input in the form of text, images, and audio.  Each
   modality is mapped to a feature representation. Feature
   representations are fused and passed to an LSTM which predicts
   whether a perpetrator is mentioned (label $l=1$) or not ($l=0$).}
 \label{fig-overview}
\end{figure}

\begin{figure}[t]
 \includegraphics[width=0.47\textwidth]{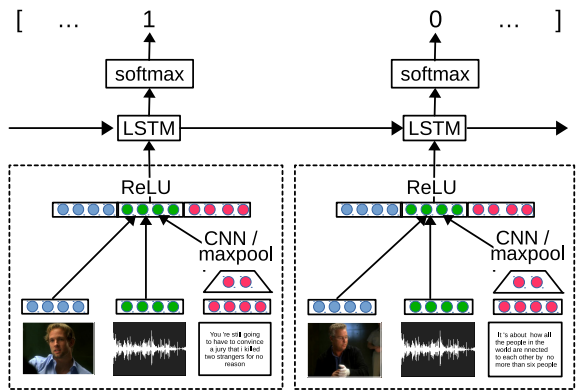}
\vspace*{-.2cm}
\caption{Illustration of input/output structure of our LSTM model for
  two time steps.}
 \label{fig-model}
\end{figure} 

A sketch of our inference task is shown in Figure~\ref{fig-overview}.
The core of our model (see Figure~\ref{fig-model}) is a
one-directional long-short term memory network (LSTM;
\newcite{Hochreiter:1997,Zaremba:2014}).  LSTM cells are a variant of
recurrent neural networks with a more complex computational unit which
have emerged as a popular architecture due to their representational
power and effectiveness at capturing long-term dependencies.  LSTMs
provide ways to selectively store and forget aspects of previously
seen inputs, and as a consequence can memorize information over longer
time periods. Through input, output and forget gates, they can
flexibly regulate the extent to which inputs are stored, used, and
forgotten.

The LSTM processes a sequence of (possibly multi-modal) inputs
$s=\{x^h_1, x^h_2, ..., x^h_N\}$. It utilizes a memory slot $c_t$ and
a hidden state $h_t$ which are incrementally updated at each time
step~$t$. Given input $x_t$, the previous latent state $h_{t-1}$ and
previous memory state~$c_{t-1}$, the latent state~$h_t$ for time~$t$
and the updated memory state~$c_t$, are computed as follows:
\begin{align*}
&\begin{bmatrix}i_t\\f_t\\o_t\\\hat{c_t}\end{bmatrix} =
  \begin{bmatrix}\sigma\\\sigma\\\sigma\\\tanh\end{bmatrix}
  W\begin{bmatrix}h_{t-1}\\x_t\end{bmatrix}\\
&c_t = f_t \odot c_{t-1} + i_t \odot \hat{c_t}\\
&h_t = o_t \odot \tanh(c_t).
\end{align*}
The weight matrix $W$ is estimated during inference, and $i$, $o$, and
$f$ are memory gates.

As mentioned earlier, the input to our model consists of a sequence of
sentences, either spoken utterances or scene descriptions (we do not
use speaker information). We further augment textual input with
multi-modal information obtained from the alignment of screenplays to
video (see Section~\ref{sec:csi}).

\paragraph{Textual modality} Words in each sentence are mapped to
50-dimensional GloVe embeddings, pre-trained on Wikipedia and
Gigaword~\cite{Pennington:2014}.  Word embeddings are subsequently
concatenated and padded to the maximum sentence length observed in our
data set in order to obtain fixed-length input vectors. The resulting
vector is passed through a convolutional layer with max-pooling to
obtain a sentence-level representation~$x^s$. Word embeddings are
fine-tuned during training.

\paragraph{Visual modality} We obtain the video corresponding to the
time span covered by each sentence and sample one frame per sentence
from the center of the associated period.\footnote{We also
  experimented with multiple frames per sentence but did not observe
  any improvement in performance.} We then map each frame to a
1,536-dimensional visual feature vector~$x^v$ using the final hidden
layer of a pre-trained convolutional network which was optimized for
object classification~(inception-v4;~\newcite{Szegedy:2016}).


\paragraph{Acoustic modality} For each sentence, we extract the audio
track from the video which includes all sounds and background music
but no spoken dialog. We then obtain Mel-frequency cepstral
coefficient (MFCC) features from the continuous signal. MFCC features
were originally developed in the context of speech
recognition~\cite{Davis:1990,Sahidullah:2012}, but have also been
shown to work well for more general sound
classification~\cite{Chachada:2014}. We extract a 13-dimensional MFCC
feature vector for every five milliseconds in the video. For each
input sentence, we sample five MFCC feature vectors from its
associated time interval, and concatenate them in chronological order
into the acoustic input~$x^a$.\footnote{Preliminary experiments showed
  that concatenation outperforms averaging or relying on a single
  feature vector.}

\paragraph{Modality Fusion}

Our model learns to fuse multi-modal input as part of its overall
architecture. We use a general method to obtain any combination of
input modalities (i.e.,~not necessarily all three).  Single modality
inputs are concatenated into an \mbox{$m$-dimensional} vector (where
$m$~is the sum of dimensionalities of all the input modalities).  We
then multiply this vector with a weight matrix $W^h$ of dimension~$m
\times n$, add an $m$-dimensional bias $b^h$, and pass the result through
a rectified linear unit (ReLU):
\begin{align*}
 x^h = \mathrm{ReLU}([x^s;x^v;x^a] W^h+b^h)
\end{align*}
The resulting multi-modal representation~$x^h$ is of dimension~$n$ and
passed to the LSTM (see Figure~\ref{fig-model}).


\section{Evaluation}
\label{sec:evaluation}

In our experiments we investigate what type of {\it knowledge} and
{\it strategy} are necessary for identifying the perpetrator in a CSI
episode. In order to shed light on the former question we compare
variants of our model with access to information from different
modalities. We examine different inference strategies by comparing the
LSTM to three baselines. The first one lacks the ability to flexibly
fuse multi-modal information (a CRF), while the second one does not
have a notion of history, classifying inputs independently (a
multilayer perceptron). Our third baseline is a rule-base system that
neither uses multi-modal inputs nor has a notion of history.  We also
compare the LSTM to humans watching CSI. Before we report our results,
we describe our setup and comparison models in more detail.

\subsection{Experimental Settings}
\label{sec-setup}

Our CSI data consists of 39~episodes giving rise to 59~cases (see
Table~\ref{tab-data-stats}). The model was trained on 53~cases using
cross-validation (five splits with 47/6 training/test cases). The remaining 
6~cases were used as truly held-out test data for final evaluation.

We trained our model using ADAM with stochastic gradient-descent and
mini-batches of six episodes. Weights were initialized randomly,
except for word embeddings which were initialized with pre-trained
$50$-dimensional GloVe vectors~\cite{Pennington:2014}, and fine-tuned
during training. We trained our networks for 100 epochs and report the
best result obtained during training. All results are averages of five
runs of the network. Parameters were optimized using two
cross-validation splits.

The sentence convolution layer has three filters of sizes $3, 4, 5$
each of which after convolution returns $75$-dimensional output. The
final sentence representation~$x^s$ is obtained by concatenating the
output of the three filters and is of dimension~$225$. We set the size
of the hidden representation of merged cross-modal inputs~$x^h$
to~$300$. The LSTM has one layer with~$128$ nodes. We set the learning
rate to~$0.001$ and apply dropout with probability of~$0.5$.

We compared model output against the gold standard of perpetrator
mentions which we collected as part of our annotation effort (second
pass).



\subsection{Model Comparison}

\paragraph{CRF} Conditional Random Fields \cite{Lafferty:2001} are
probabilistic graphical models for sequence labeling. The comparison
allows us to examine whether the LSTM's use of long-term memory and
(non-linear) feature integration is beneficial for sequence
prediction. We experimented with a variety of features for the CRF,
and obtained best results when the input sentence is represented by
concatenated word embeddings.

\paragraph{MLP} We also compared the LSTM against a multi-layer
perceptron with two hidden layers, and a softmax output layer. We
replaced the LSTM in our overall network structure with the MLP,
keeping the methodology for sentence convolution and modality fusion
and all associated parameters fixed to the values described in
Section~\ref{sec-setup}. The hidden layers of the MLP have ReLU
activations and layer-size of~128, as in the LSTM. We set the learning
rate to~$0.0001$. The MLP makes independent predictions for each
element in the sequence. This comparison sheds light on the importance
of {\it sequential} information for perpetrator identification
task. All results are best checkpoints over 100 training epochs,
averaged over five runs.

\paragraph{PRO} Aside from the supervised models described so far, we
developed a simple rule-based system which does not require access to
labeled data. The system defaults to the perpetrator class for any
sentence containing a personal (e.g., \textsl{you}), possessive
(e.g.,~\textsl{mine}) or reflexive pronoun
(e.g.,\textsl{~ourselves}). In other words, it assumes that every
pronoun refers to the perpetrator. Pronoun mentions were identified
using string-matching and a precompiled list of~31 pronouns. This
system cannot incorporate any acoustic or visual data.
   
\paragraph{Human Upper Bound} Finally, we compared model performance
against humans. In our annotation task
(Section~\ref{ssec-annotation-behavior}), participants annotate
sentences incrementally, while watching an episode for the first time.
The annotations express their \emph{belief} as to whether the
perpetrator is mentioned. We evaluate these first-pass guesses against
the gold standard (obtained in the second-pass annotation).

 \begin{table}[t]
\setlength\tabcolsep{5pt}
 \begin{tabular}{|@{~}l@{\hspace{-.1cm}}ccc@{~}|@{~}c@{~~}c@{~~}c@{~}|@{~}c@{~~}c@{~~}c@{~}|}
\hline
{\bf Model}   & \multicolumn{3}{@{~}c@{~}}{\bf Modality} &
                                                     \multicolumn{3}{c}{\bf Cross-val}&\multicolumn{3}{c|}{\bf Held-out} \\
 & {\bf T} & {\bf V} & {\bf A}  & {\bf pr} & {\bf re} & {\bf f1}
                                      & {\bf pr} & {\bf re} & {\bf f1}
   \\\hline
                 PRO    &+& -- & --& 19.3 & 76.3 & 31.6 & 19.5 & 77.2
                                                      & 31.1 \\     
                 CRF    &+&--&--& 33.1 & 15.4 & 20.5&30.2&16.1&21.0\\\hline\hline
\multirow{4}{*}{MLP}    &+&--&--& 36.7 & 32.5 & 33.7&35.9&36.8&36.3\\
                        &+&+&--& 37.4 & 35.1 & 35.1&38.0&41.0&39.3\\ 
                        &+&--&+& 39.6 & 34.2 & 35.7&38.7&36.5&37.5\\ 
                        &+&+&+& 38.4 & 34.6 & 35.7&38.5&42.3&40.2\\ \hline\hline
\multirow{4}{*}{LSTM}   &+&--&--& 39.2 & 45.7 & 41.3&36.9&50.4&42.3 \\
                        &+&+&--& 39.9 & 48.3 & 43.1&40.9&54.9&46.8 \\
                        &+&--&+& 39.2 & 52.0 & 44.0&36.8&56.3&44.5\\
                        &+&+&+& 40.6 & 49.7 &
                                              44.1&42.8&51.2&46.6\\\hline\hline
                  Humans & & & & 74.1 & 49.4 & 58.2&76.3&60.2&67.3 \\\hline

 \end{tabular}
 \setlength\tabcolsep{6pt}
 \caption{Precision (pr) recall (re) and f1 for detecting the minority
   class (perpetrator mentioned) for humans (bottom) and various
   systems. We report results with
   cross-validation  (center) and on a held-out data set
   (right) using the textual (T) visual (V), and auditory (A) modalities.} 
 \label{tab-results}
 \end{table}

\begin{figure*}[t]
\begin{center}
\hspace*{-.2cm}\includegraphics[scale=.75]{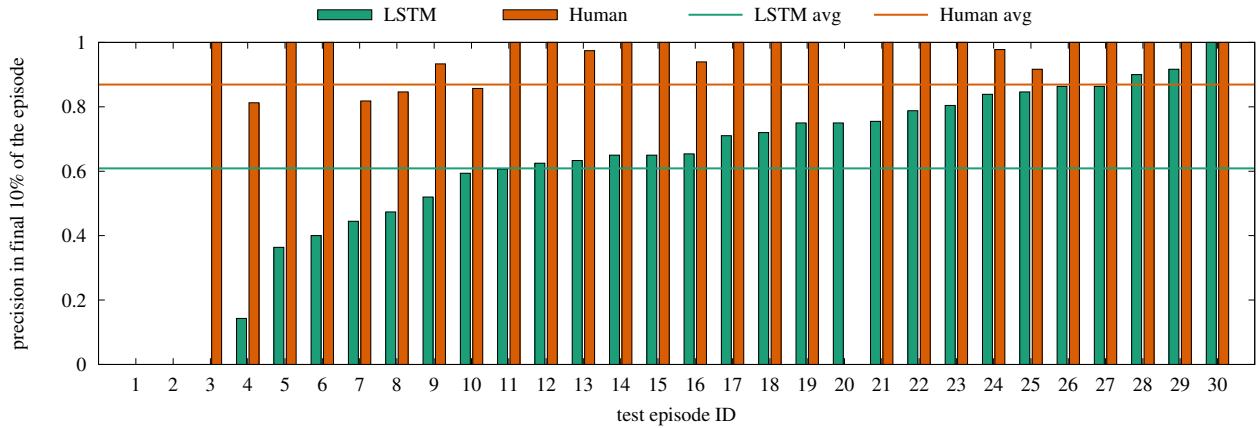}
\end{center}
\caption{
  Precision in the final~10\% of an episode, for 30~test episodes from
  five cross-validation splits. We show scores per episode and global
  averages (horizontal bars). Episodes are ordered by increasing model
  precision.}
\label{fig-when-guessed}
\end{figure*}

\subsection{Which Model Is the Best Detective?}
\label{sec-eval1}

We report precision, recall and f1 on the minority class, focusing on
how accurately the models identify perpetrator mentions.
Table~\ref{tab-results} summarizes our results, averaged across five
cross-validation splits (left) and on the truly held-out test episodes
(right).

Overall, we observe that humans outperform all comparison models. In
particular, human precision is superior, whereas recall is comparable,
with the exception of~PRO which has high recall (at the expense of
precision) since it assumes that all pronouns refer to
perpetrators. We analyze the differences between model and human
behavior in more detail in Section~\ref{sec-eval-3}. With regard to
the LSTM, both visual and acoustic modalities bring improvements over
the textual modality, however, their contribution appears to be
complementary. We also experimented with acoustic and visual features
on their own, but without high-level textual information, the LSTM
converges towards predicting the majority class only. Results on the
held-out test set reveal that our model generalizes well to unseen
episodes, despite being trained on a relatively small data sample
compared to standards in deep learning.



The LSTM consistently outperforms the non-incremental MLP. This shows
that the ability to utilize information from previous inputs is
essential for this task. This is intuitively plausible; in order to
identify the perpetrator, viewers must be aware of the plot's
development and make inferences while the episode evolves. The CRF is
outperformed by all other systems, including rule-based PRO. In
contrast to the MLP and PRO, the CRF utilizes sequential information,
but cannot flexibly fuse information from different modalities or
exploit non-linear mappings like neural models. The only type of input
which enabled the CRF to predict perpetrator mentions were
concatenated word embeddings (see Table~\ref{tab-results}). We trained
CRFs on audio or visual features, together with word embeddings,
however these models converged to only predicting the majority
class. This suggests that CRFs do not have the capacity to model
complex long sequences and draw meaningful inferences based on them.
PRO achieves a reasonable f1 score but does so because it achieves high 
recall at the expense of very low precision. The precision-recall tradeoff
is much more balanced for the neural systems.

\begin{figure*}[ht!]
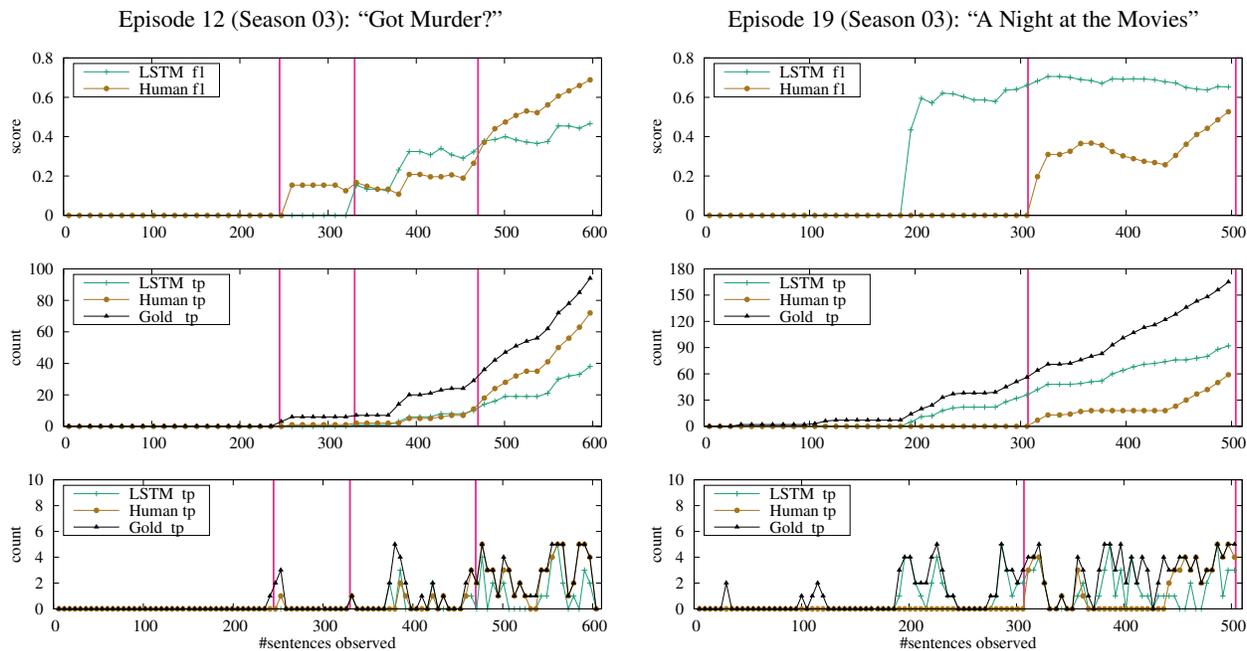

\hspace*{-.3cm}\begin{tabular}{l@{~}l}
\multicolumn{1}{c}{\footnotesize{Episode~12 (Season 03): ``Got Murder?''}} &
                                                \multicolumn{1}{c}{\footnotesize{Episode~19
                                                                           (Season~03):
                                                                           ``A
                                                                           Night
                                                                           at
                                                                           the
                                                                           Movies''}}
  \\
 \includegraphics[width=0.51\textwidth]{\detokenize{./pictures/split3_1_vs_human_fewer}}
  & 
 \includegraphics[width=0.51\textwidth]{\detokenize{./pictures/split5_2_vs_human_fewer}}
\end{tabular}
\vspace*{-.5cm}
\caption{Human and LSTM behavior over the course of two episodes (left
  and right). Top plots show cumulative f1; true positives
  (tp) are shown cumulatively (center) and as individual counts for
  each interval (bottom).  Statistics relating to {\it gold}
  perpetrator mentions are shown in black. Red vertical bars show when
  humans press the red button to indicate that they (think they) have
  identified the perpetrator.}
  \label{fig-performance}
\end{figure*}

\begin{table*}[t]
\begin{tabular}{l@{~}l}
\begin{minipage}[t]{3.8in}
\renewcommand{\tablename}{Figure}
\setcounter{table}{7}

\hspace*{-.2cm}\includegraphics[width=1.06\textwidth]{\detokenize{./pictures/scatter_freq_cumul}}
\caption{Number of times the red button is pressed by each annotator
  individually (bottom) and by all three within each time interval and
  cumulatively (top). Times are normalized with respect to
  length. Statistics are averaged across 18/12/9 cases per annotator~1/2/3.}
 \label{red-button}
\end{minipage} &
\renewcommand{\tablename}{Table}
\setcounter{table}{2}
\raisebox{1.7cm}[0pt]{\begin{minipage}[t]{2.3in}
\begin{small}
\begin{center}
  \begin{tabular}{|lrrr|}
  \hline
          \multicolumn{4}{|c|}{First correct perpetrator prediction}\\
          & min & max & avg \\\hline
    LSTM & 2   & 554 & 141 \\
    Human & 12 & 1014& 423 \\\hline
  \end{tabular}
\end{center}
\end{small}
\caption{Sentence ID in the script where the LSTM and Humans predict
  the true perpetrator for the first time. We show the earliest (min)
  latest (max) and average (avg) prediction time over 30 test episodes
  (five cross-validation splits).\label{tab:firsttrueguesses}}
\end{minipage}}
\end{tabular}
\end{table*}
\subsection{Can the Model Identify the Perpetrator?}


In this section we assess more directly how the LSTM compares against
humans when asked to identify the perpetrator by the end of a CSI
episode. Specifically, we measure precision in the final~10\% of an
episode, and compare human performance (first-pass guesses) and an
LSTM model which uses all three
modalities. Figure~\ref{fig-when-guessed} shows precision results for
30~test episodes (across five cross-validation splits) and average
precision as horizontal bars.

\begin{table*}[t]
\begin{footnotesize}
\begin{tabular}{|p{1.5cm}p{1.1cm}p{1.3cm}p{1.1cm}p{1.6cm}p{1.9cm}p{1.2cm}p{1.2cm}p{1.5cm}}
\multicolumn{8}{c}{Episode 03 (Season 03): ``Let the Seller Beware''}
  \\ \hline
 \centering s1 &\centering s2 &\centering s3&\centering s4&\centering s5&\centering s6&\centering s7&\centering s8& \multicolumn{1}{c|}{s9}\\
\hline
\cellcolor{leaRed!00.9}{\it Grissom pulls out a small evidence bag with the filling}
&\cellcolor{leaRed!09.9}{\it He puts it on the table}
&\cellcolor{leaRed!18.1}{\it Tooth filling 0857}
&\cellcolor{leaRed!55.2}{\it 10-7-02}
&\cellcolor{leaRed!84.6}{\bf Brass} We also found \textcolor{leaGreen}{\bf your} fingerprints and \textcolor{leaGreen}{\bf your} hair
&\cellcolor{leaRed!95.5}{\bf Peter B.} Look \textcolor{leaGreen}{\bf I}'m sure you'll find \textcolor{leaGreen}{\bf me} all over the house
&\cellcolor{leaRed!74.2} {\bf Peter B.} \textcolor{leaGreen}{\bf I} wanted to buy it
&\cellcolor{leaRed!86.8} {\bf Peter B.} \textcolor{leaGreen}{\bf I} was everywhere
&\cellcolor{leaRed!96.4}{\bf Brass} well \textcolor{leaGreen}{\bf you} made sure \textcolor{leaGreen}{\bf you} were everywhere too didn't \textcolor{leaGreen}{\bf you}?\\\hline
\end{tabular}
\vspace{2ex}

\begin{tabular}{|p{1.4cm}p{1.8cm}p{1.7cm}p{1.9cm}p{1.3cm}p{1.3cm}p{1.7cm}p{1.7cm}|}
\multicolumn{8}{c}{Episode 21 (Season 05): ``Committed''}
  \\  \hline
\centering s1 &\centering s2 &\centering s3&\centering s4&\centering s5&\centering s6&\centering s7&\multicolumn{1}{c|}{s8} \\\hline
\cellcolor{leaRed!1.2}{\bf Grissom} What's so amusing?
&\cellcolor{leaRed!60.7}{\bf Adam Trent} So let's say you find out who did it and maybe it's me.
&\cellcolor{leaRed!41}{\bf Adam Trent} What are you going to do?
&\cellcolor{leaRed!90}{\bf Adam Trent} Are you going to convict me of murder and put me in a bad place?
&\cellcolor{leaRed!91}{\it Adam smirks and starts biting his nails.}
&\cellcolor{leaRed!25}{\bf Grissom} Is it you?
&\cellcolor{leaRed!29}{\bf Adam Trent} Check the files sir.
&\cellcolor{leaRed!93}{\bf Adam Trent} I'm a rapist not a murderer.\\\hline
\end{tabular}\end{footnotesize}
\caption{Excerpts of CSI episodes together with model predictions. Model
  confidence ($p(l=1)$) is illustrated in red, with 
  darker shades corresponding to higher confidence. True perpetrator
  mentions are highlighted in blue.  Top: a conversation
  involving the true perpetrator.  Bottom: a conversation with a
  suspect who is not the perpetrator. }
\label{tab-qualitative}
\end{table*}


Perhaps unsurprisingly, human performance is superior, however, the
model achieves an average precision of 60\% which is encouraging
(compared to 85\% achieved by humans).  Our results also show a
moderate correlation between model and humans: episodes which are
difficult for the LSTM (see left side of the plot in
Figure~\ref{fig-when-guessed}) also result in lower human
precision. Two episodes on the very left of the plot have~0\%
precision and are special cases. The first one revolves around a
suicide, which is not strictly speaking a crime, while the second one
does not mention the perpetrator in the final~10\%.

\subsection{How Is the Model Guessing?}
\label{sec-eval-3}

We next analyze how the model's guessing ability compares to humans.
Figure~\ref{fig-performance} tracks model behavior over the course of
two episodes, across 100~equally sized intervals. We show the
cumulative development of f1 (top plot), cumulative true positive
counts (center plot), and true positive counts within each interval
(bottom plot). Red bars indicate times at which annotators pressed the red button.

Figure~\ref{fig-performance} (right) shows that humans may outperform the LSTM in precision (but
not necessarily in recall).  Humans are more cautious at guessing
the perpetrator: the first human guess appears around sentence 300
(see the leftmost red vertical bars in Figure~\ref{fig-performance} right),
the first model guess around sentence 190, and the first true mention
around sentence 30. Once humans guess the perpetrator, however, they
are very precise and consistent. Interestingly, model guesses at the
start of the episode closely follow the pattern of gold-perpetrator
mentions (bottom plots in Figure~\ref{fig-performance}). This
indicates that early model guesses are not noise, but meaningful
predictions.


Further analysis of human responses is illustrated in
Figure~\ref{red-button}. For each of our three annotators we plot the
points in each episode where they press the red button to indicate
that they know the perpetrator (bottom). We also show the number of
times (all three) annotators pressed the red button individually for
each interval and cumulatively over the course of the episode. Our
analysis reveals that viewers tend to press the red button more
towards the end, which is not unexpected since episodes are inherently
designed to obfuscate the identification of the perpetrator. Moreover,
Figure~\ref{red-button} suggests that there are two types of viewers:
\emph{eager} viewers who like our model guess early on, change their
mind often, and therefore press the red button frequently (annotator 1
pressed the red button 6.1 times on average per episode) and
\emph{conservative} viewers who guess only late and press the red
button less frequently (on average annotator 2 pressed the red button
2.9 times per episode, and annotator~3 and 3.7 times).  Notice that
statistics in Figure~\ref{red-button} are averages across several
episodes each annotator watched and thus viewer behavior is unlikely
to be an artifact of individual episodes (e.g.,~featuring more or less
suspects).  Table~3 provides further evidence that the LSTM behaves
more like an eager viewer.  It presents the time in the episode (by
sentence count) where the model correctly identifies the perpetrator
for the first time. As can be seen, the minimum and average
identification times are lower for the LSTM compared to human viewers.

Table~\ref{tab-qualitative} shows model predictions on two CSI
screenplay excerpts.  We illustrate the degree of the model's belief
in a perpetrator being mentioned by color intensity. True perpetrator
mentions are highlighted in blue. In the first example, the model
mostly identifies perpetrator mentions correctly. In the second
example, it identifies seemingly plausible sentences which, however,
refer to a suspect and not the true perpetrator.

\subsection{What if There Is No Perpetrator?}
\label{sec:what-if-there}

\begin{figure}[t]
 \hspace{-1.2ex}\includegraphics[width=0.512\textwidth]{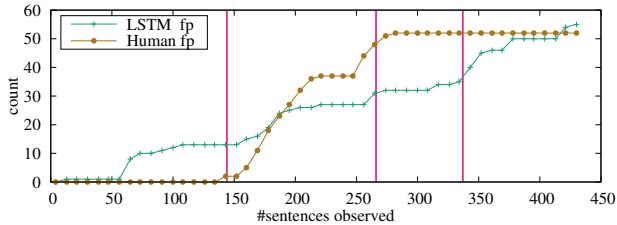} 
 \caption{Cumulative counts of false positives (fp) for the LSTM and a
   human viewer for an episode with no perpetrator (the victim
   committed suicide). Red vertical bars show the times at which the
   viewer pressed the red button indicating that they (think they)
   have identified the perpetrator.}
\label{fig-suicide}
\end{figure} 

In our experiments, we trained our model on CSI episodes which
typically involve a crime, committed by a perpetrator, who is
ultimately identified. How does the LSTM generalize to episodes {\it
  without} a crime, e.g.,~because the ``victim'' turns out to have
committed suicide? To investigate how model and humans alike respond to
atypical input we present both with an episode featuring a suicide,
i.e.,~an episode which did not have any true positive perpetrator
mentions.

Figure~\ref{fig-suicide} tracks the incremental behavior of a human
viewer and the model while watching the suicide episode.  Both are
primed by their experience with CSI episodes to identify characters in
the plot as potential perpetrators, and predict false positive perpetrator mentions. The human realizes after roughly
two thirds of the episode that there is no perpetrator involved (he
does not annotate any subsequent sentences as ``perpetrator
mentioned''), whereas the LSTM continues to make perpetrator predictions
until the end of the episode. The LSTM's behavior is presumably an
artifact of the recurring pattern of discussing the perpetrator in the
very end of an episode.

\section{Conclusions}
\label{sec:conclusions}

In this paper we argued that crime drama is an ideal testbed for
models of natural language understanding and their ability to draw
inferences from complex, multi-modal data. The inference task is
well-defined and relatively constrained: every episode poses and
answers the same ``whodunnit'' question. We have formalized
perpetrator identification as a sequence labeling problem and
developed an LSTM-based model which learns incrementally from complex
naturalistic data. We showed that \emph{multi-modal} input is
essential for our task as well an \emph{incremental} inference
strategy with flexible access to previously observed
information. Compared to our model, humans guess cautiously in the
beginning, but are consistent in their predictions once they have a
strong suspicion. The LSTM starts guessing earlier, leading to
superior initial true-positive rates, however, at the cost of
consistency.

There are many directions for future work. Beyond perpetrators, we may
consider how suspects emerge and disappear in the course of an
episode. Note that we have obtained suspect annotations but did not
used them in our experiments. It should also be interesting to examine
how the model behaves out-of-domain, i.e.,~when tested on other crime
series, e.g.,~``Law and Order''. Finally, more detailed analysis of
what happens in an episode (e.g., what actions are performed, by who,
when, and where) will give rise to deeper understanding enabling
applications like video summarization and skimming.

\paragraph{Acknowledgments} 
The authors gratefully acknowledge the support of the European
Research Council (award number 681760; Frermann, Lapata) and H2020 EU
project SUMMA (award number 688139/H2020-ICT-2015; Cohen). We also
thank our annotators, the anonymous TACL reviewers whose feedback
helped improve the present paper, and members of EdinburghNLP for
helpful discussions and suggestions.

\bibliography{csi} 

\begin{thebibliography}{}

\bibitem[\protect\citename{Antol \bgroup et al.\egroup }2015]{antol2015vqa}
Stanislaw Antol, Aishwarya Agrawal, Jiasen Lu, \~Margaret Mitchell, Dhruv
  Batra, C.~Lawrence~Zitnick, and Devi Parikh.
\newblock 2015.
\newblock {VQA: Visual Question Answering}.
\newblock In {\em Proceedings of the IEEE International Conference on Computer
  Vision (ICCV)}, pages 2425--2433, Santiago, Chile.

\bibitem[\protect\citename{Bojanowski \bgroup et al.\egroup
  }2013]{Bojanowski:ea:2013}
Piotr Bojanowski, Francis Bach, Ivan Laptev, Jean Ponce, Cordelia Schmid, and
  Josef Sivic.
\newblock 2013.
\newblock Finding actors and actions in movies.
\newblock In {\em The IEEE International Conference on Computer Vision (ICCV)},
  pages 2280--2287, Sydney, Australia.

\bibitem[\protect\citename{Boreczky and Wilcox}1998]{Boreczky:1998}
John~S. Boreczky and Lynn~D. Wilcox.
\newblock 1998.
\newblock A hidden {Markov} model framework for video segmentation using audio
  and image features.
\newblock In {\em Proceedings of the 1998 {IEEE} International Conference on
  Acoustics, Speech and Signal Processing (ICASSP)}, pages 3741--3744, Seattle,
  Washington, USA.

\bibitem[\protect\citename{Bowman \bgroup et al.\egroup
  }2015]{bowman-EtAl:2015:EMNLP}
Samuel~R. Bowman, Gabor Angeli, Christopher Potts, and Christopher~D. Manning.
\newblock 2015.
\newblock A large annotated corpus for learning natural language inference.
\newblock In {\em Proceedings of the 2015 Conference on Empirical Methods in
  Natural Language Processing}, pages 632--642, Lisbon, Portugal.

\bibitem[\protect\citename{Bruni \bgroup et al.\egroup }2014]{Bruni:2014}
Elia Bruni, Nam~Khanh Tran, and Marco Baroni.
\newblock 2014.
\newblock Multimodal distributional semantics.
\newblock {\em J. Artif. Int. Res.}, 49(1):1--47, January.

\bibitem[\protect\citename{Chachada and Kuo}2014]{Chachada:2014}
Sachin Chachada and C.-C.~Jay Kuo.
\newblock 2014.
\newblock Environmental sound recognition: A survey.
\newblock {\em APSIPA Transactions on Signal and Information Processing}, 3.

\bibitem[\protect\citename{Cohen}1960]{Cohen:1960}
Jacob Cohen.
\newblock 1960.
\newblock A coefficient of agreement for nominal scales.
\newblock {\em Educational and Psychological Measurement}, 20(1):37--46.

\bibitem[\protect\citename{Cour \bgroup et al.\egroup }2008]{Cour:ea:2008}
Timothee Cour, Chris Jordan, Eleni Miltsakaki, and Ben Taskar.
\newblock 2008.
\newblock Movie/script: Alignment and parsing of video and text transcription.
\newblock In {\em Proceedings of the 10th European Conference on Computer
  Vision}, pages 158--171, Marseille, France.

\bibitem[\protect\citename{Davis and Mermelstein}1990]{Davis:1990}
Steven~B. Davis and Paul Mermelstein.
\newblock 1990.
\newblock Comparison of parametric representations for monosyllabic word
  recognition in continuously spoken sentences.
\newblock In Alex Waibel and Kai-Fu Lee, editors, {\em Readings in Speech
  Recognition}, pages 65--74. Morgan Kaufmann Publishers Inc., San Francisco,
  California, USA.

\bibitem[\protect\citename{Dimitrova \bgroup et al.\egroup
  }2000]{Dimitrova:2000}
Nevenka Dimitrova, Lalitha Agnihotri, and Gang Wei.
\newblock 2000.
\newblock Video classification based on {HMM} using text and faces.
\newblock In {\em Proceedings of the 10th European Signal Processing Conference
  (EUSIPCO)}, pages 1--4. IEEE.

\bibitem[\protect\citename{Elliott and Keller}2013]{elliott2013image}
Desmond Elliott and Frank Keller.
\newblock 2013.
\newblock Image description using visual dependency representations.
\newblock In {\em Proceedings of the 2013 Conference on Empirical Methods in
  Natural Language Processing}, pages 1292--1302, Seattle, Washington, USA.

\bibitem[\protect\citename{Gorinski and Lapata}2015]{Gorinski:2015}
Philip~John Gorinski and Mirella Lapata.
\newblock 2015.
\newblock Movie script summarization as graph-based scene extraction.
\newblock In {\em Proceedings of the 2015 Conference of the North American
  Chapter of the Association for Computational Linguistics: Human Language
  Technologies}, pages 1066--1076, Denver, Colorado, USA.

\bibitem[\protect\citename{Hermann \bgroup et al.\egroup
  }2015]{Hermann:ea:2015}
Karl~Moritz Hermann, Tomas Kocisky, Edward Grefenstette, Lasse Espeholt, Will
  Kay, Mustafa Suleyman, and Phil Blunsom.
\newblock 2015.
\newblock Teaching machines to read and comprehend.
\newblock In C.~Cortes, N.~D. Lawrence, D.~D. Lee, M.~Sugiyama, and R.~Garnett,
  editors, {\em Advances in Neural Information Processing Systems 28}, pages
  1693--1701. Curran Associates, Inc.

\bibitem[\protect\citename{Hill \bgroup et al.\egroup }2015]{Hill:ea:2015}
Felix Hill, Anoine Bordes, Sumit Chopra, and Jason Weston.
\newblock 2015.
\newblock The {Goldilocks} principle: Reading children's books with explicit
  memory representations.
\newblock In {\em Proceedings of the 3rd International Conference on Learning
  Representations (ICLR)}, San Diego, California, USA.

\bibitem[\protect\citename{Hochreiter and Schmidhuber}1997]{Hochreiter:1997}
Sepp Hochreiter and J\"{u}rgen Schmidhuber.
\newblock 1997.
\newblock Long short-term memory.
\newblock {\em Neural Computation}, 9(8):1735--1780, November.

\bibitem[\protect\citename{Johnson \bgroup et al.\egroup
  }2015]{johnson2015image}
Justin Johnson, Ranjay Krishna, Michael Stark, Li-Jia Li, David~A Shamma,
  Michael~S Bernstein, and Li~Fei-Fei.
\newblock 2015.
\newblock Image retrieval using scene graphs.
\newblock In {\em Proceedings of the 2015 IEEE Conference on Computer Vision
  and Pattern Recognition (CVPR)}, pages 3668--3678, Boston, Massachusetts,
  USA.

\bibitem[\protect\citename{Karpathy and Fei-Fei}2015]{karpathy2015deep}
Andrej Karpathy and Li~Fei-Fei.
\newblock 2015.
\newblock Deep visual-semantic alignments for generating image descriptions.
\newblock In {\em Proceedings of the IEEE Conference on Computer Vision and
  Pattern Recognition}, pages 3128--3137, Boston, Massachusetts.

\bibitem[\protect\citename{Kiela and Bottou}2014]{kiela-bottou:2014:EMNLP2014}
Douwe Kiela and L\'{e}on Bottou.
\newblock 2014.
\newblock Learning image embeddings using convolutional neural networks for
  improved multi-modal semantics.
\newblock In {\em Proceedings of the 2014 Conference on Empirical Methods in
  Natural Language Processing (EMNLP)}, pages 36--45, Doha, Qatar.

\bibitem[\protect\citename{Lafferty \bgroup et al.\egroup }2001]{Lafferty:2001}
John~D. Lafferty, Andrew McCallum, and Fernando C.~N. Pereira.
\newblock 2001.
\newblock Conditional random fields: Probabilistic models for segmenting and
  labeling sequence data.
\newblock In {\em Proceedings of the 18th International Conference on Machine
  Learning}, pages 282--289, San Francisco, CA, USA. Morgan Kaufmann Publishers
  Inc.

\bibitem[\protect\citename{Lazaridou \bgroup et al.\egroup
  }2015]{lazaridou-pham-baroni:2015:NAACL-HLT}
Angeliki Lazaridou, Nghia~The Pham, and Marco Baroni.
\newblock 2015.
\newblock Combining language and vision with a multimodal skip-gram model.
\newblock In {\em Proceedings of the 2015 Conference of the North American
  Chapter of the Association for Computational Linguistics: Human Language
  Technologies}, pages 153--163, Denver, Colorado, USA.

\bibitem[\protect\citename{Lin \bgroup et al.\egroup }2014]{Lin:ea:2014}
Dahua Lin, Sanja Fidler, Chen Kong, and Raquel Urtasun.
\newblock 2014.
\newblock Visual semantic search: Retrieving videos via complex textual
  queries.
\newblock In {\em IEEE Conference on Computer Vision and Pattern Recognition},
  pages 2657--2664, Columbus, Ohio, USA.

\bibitem[\protect\citename{Myers and Rabiner}1981]{Myers:ea:1981}
Cory~S. Myers and Lawrence~R. Rabiner.
\newblock 1981.
\newblock A comparative study of several dynamic time-warping algorithms for
  connected word recognition.
\newblock {\em The Bell System Technical Journal}, 60(7):1389--1409.

\bibitem[\protect\citename{Naphide and Huang}2001]{Naphide:2001}
Milind~R. Naphide and Thomas~S. Huang.
\newblock 2001.
\newblock A probabilistic framework for semantic video indexing, filtering, and
  retrieval.
\newblock {\em IEEE Transactions on Multimedia}, 3(1):141--151.

\bibitem[\protect\citename{Ortiz \bgroup et al.\egroup
  }2015]{ortiz2015learning}
Luis Gilberto~Mateos Ortiz, Clemens Wolff, and Mirella Lapata.
\newblock 2015.
\newblock Learning to interpret and describe abstract scenes.
\newblock In {\em Proceedings of the 2015 NAACL: Human Language Technologies},
  pages 1505--1515, Denver, Colorado, USA.

\bibitem[\protect\citename{Pennington \bgroup et al.\egroup
  }2014]{Pennington:2014}
Jeffrey Pennington, Richard Socher, and Christopher~D. Manning.
\newblock 2014.
\newblock {GloVe}: Global vectors for word representation.
\newblock In {\em Proceedings of the 2014 Conference on Empirical Methods in
  Natural Language Processing (EMNLP)}, pages 1532--1543, Doha, Qatar.

\bibitem[\protect\citename{Rajpurkar \bgroup et al.\egroup
  }2016]{rajpurkar-EtAl:2016:EMNLP2016}
Pranav Rajpurkar, Jian Zhang, Konstantin Lopyrev, and Percy Liang.
\newblock 2016.
\newblock {SQuAD}: 100,000+ questions for machine comprehension of text.
\newblock In {\em Proceedings of the 2016 Conference on Empirical Methods in
  Natural Language Processing}, pages 2383--2392, Austin, Texas, USA.

\bibitem[\protect\citename{Rasheed \bgroup et al.\egroup }2005]{Rasheed:2005}
Zeeshan Rasheed, Yaser Sheikh, and Mubarak Shah.
\newblock 2005.
\newblock {On the use of computable features for film classification}.
\newblock {\em IEEE Transactions on Circuits and Systems for Video Technology},
  15(1):52--64.

\bibitem[\protect\citename{Richardson \bgroup et al.\egroup
  }2013]{richardson-burges-renshaw:2013:EMNLP}
Matthew Richardson, Christopher~J.C. Burges, and Erin Renshaw.
\newblock 2013.
\newblock {MCTest}: A challenge dataset for the open-domain machine
  comprehension of text.
\newblock In {\em Proceedings of the 2013 Conference on Empirical Methods in
  Natural Language Processing}, pages 193--203, Seattle, Washington, USA.

\bibitem[\protect\citename{Rockt{\"a}schel \bgroup et al.\egroup
  }2016]{rocktaschel2016reasoning}
Tim Rockt{\"a}schel, Edward Grefenstette, Karl~Moritz Hermann, Tomas Kocisky,
  and Phil Blunsom.
\newblock 2016.
\newblock Reasoning about entailment with neural attention.
\newblock In {\em Proceedings of the 4th International Conference on Learning
  Representations (ICLR)}, San Juan, Puerto Rico.

\bibitem[\protect\citename{Rohrbach \bgroup et al.\egroup
  }2017]{Rohrbach:ea:2017}
Anna Rohrbach, Atousa Torabi, Marcus Rohrbach, Niket Tandon, Christopher Pal,
  Hugo Larochelle, Aaron Courville, and Bernt Schiele.
\newblock 2017.
\newblock Movie description.
\newblock {\em International Journal of Computer Vision}, 123(1):94--120.

\bibitem[\protect\citename{Sahidullah and Saha}2012]{Sahidullah:2012}
Md~Sahidullah and Goutam Saha.
\newblock 2012.
\newblock Design, analysis and experimental evaluation of block based
  transformation in {MFCC} computation for speaker recognition.
\newblock {\em Speech Communication}, 54(4):543--565.

\bibitem[\protect\citename{Sang and Xu}2010]{Sang:2010}
Jitao Sang and Changsheng Xu.
\newblock 2010.
\newblock Character-based movie summarization.
\newblock In {\em Proceedings of the 18th ACM International Conference on
  Multimedia}, pages 855--858, Firenze, Italy.

\bibitem[\protect\citename{Silberer \bgroup et al.\egroup
  }2016]{silberer:ea:16}
Carina Silberer, Vittorio Ferrari, and Mirella Lapata.
\newblock 2016.
\newblock Visually grounded meaning representations.
\newblock {\em IEEE Transactions on Pattern Analysis and Machine Intelligence},
  99.

\bibitem[\protect\citename{Sivic \bgroup et al.\egroup }2009]{Sivic:ea:2009}
Josef Sivic, Mark Everingham, and Andrew Zisserman.
\newblock 2009.
\newblock ``{W}ho are you?'' -- {L}earning person specific classifiers from
  video.
\newblock In {\em IEEE Conference on Computer Vision and Pattern Recognition},
  pages 1145--1152, Miami, Florida, USA.

\bibitem[\protect\citename{Sutskever \bgroup et al.\egroup
  }2014]{Sutskever:2014}
Ilya Sutskever, Oriol Vinyals, and Quoc~V. Le.
\newblock 2014.
\newblock Sequence to sequence learning with neural networks.
\newblock In {\em Proceedings of the 27th International Conference on Neural
  Information Processing Systems}, NIPS'14, pages 3104--3112, Cambridge, MA,
  USA. MIT Press.

\bibitem[\protect\citename{Szegedy \bgroup et al.\egroup }2016]{Szegedy:2016}
Christian Szegedy, Sergey Ioffe, and Vincent Vanhoucke.
\newblock 2016.
\newblock Inception-v4, inception-{ResNet} and the impact of residual
  connections on learning.
\newblock {\em CoRR}, abs/1602.07261.

\bibitem[\protect\citename{Tapaswi \bgroup et al.\egroup
  }2015]{Tapaswi:ea:2015}
Makarand Tapaswi, Martin B\"{a}uml, and Rainer Stiefelhagen.
\newblock 2015.
\newblock Aligning plot synopses to videos for story-based retrieval.
\newblock {\em International Journal of Multimedia Information Retrieval},
  (4):3--26.

\bibitem[\protect\citename{Tapaswi \bgroup et al.\egroup
  }2016]{Tapaswietal:2016}
Makarand Tapaswi, Yukun Zhu, Rainer Stiefelhagen, Antonio Torralba, Raquel
  Urtasun, and Sanja Fidler.
\newblock 2016.
\newblock {MovieQA:} {U}nderstanding stories in movies through
  question-answering.
\newblock In {\em The IEEE Conference on Computer Vision and Pattern
  Recognition (CVPR)}, pages 4631--4640, Las Vegas, Nevada.

\bibitem[\protect\citename{Venugopalan \bgroup et al.\egroup
  }2015a]{Venugopalan:2015}
Subhashini Venugopalan, Marcus Rohrbach, Jeff Donahue, Raymond~J. Mooney,
  Trevor Darrell, and Kate Saenko.
\newblock 2015a.
\newblock Sequence to sequence -- {V}ideo to text.
\newblock In {\em Proceedings of the 2015 International Conference on Computer
  Vision (ICCV)}, pages 4534--4542, Santiago, Chile.

\bibitem[\protect\citename{Venugopalan \bgroup et al.\egroup
  }2015b]{Venugopalan:2015naacl}
Subhashini Venugopalan, Huijuan Xu, Jeff Donahue, Marcus Rohrbach, Raymond
  Mooney, and Kate Saenko.
\newblock 2015b.
\newblock Translating videos to natural language using deep recurrent neural
  networks.
\newblock In {\em Proceedings the 2015 Conference of the North American Chapter
  of the Association for Computational Linguistics -- Human Language
  Technologies (NAACL HLT 2015)}, pages 1494--1504, Denver, Colorado, June.

\bibitem[\protect\citename{Vinyals \bgroup et al.\egroup
  }2015]{vinyals2015show}
Oriol Vinyals, Alexander Toshev, Samy Bengio, and Dumitru Erhan.
\newblock 2015.
\newblock Show and tell: {A} neural image caption generator.
\newblock {\em Proceedings of the 2015 IEEE Conference on Computer Vision and
  Pattern Recognition (CVPR)}, pages 3156--3164.

\bibitem[\protect\citename{Voorhees and Tice}2000]{Voorhees:Tice:2000}
Ellen~M. Voorhees and Dawn~M. Tice.
\newblock 2000.
\newblock Building a question answering test collection.
\newblock In {\em ACM Special Interest Group on Information Retrieval (SIGIR)},
  pages 200--207, Athens, Greece.

\bibitem[\protect\citename{Weston \bgroup et al.\egroup }2015]{Weston:2015}
Jason Weston, Antoine Bordes, Sumit Chopra, and Tomas Mikolov.
\newblock 2015.
\newblock Towards {AI}-complete question answering: {A} set of prerequisite toy
  tasks.
\newblock {\em CoRR}, abs/1502.05698.

\bibitem[\protect\citename{Xu \bgroup et al.\egroup }2015]{xu2015show}
Kelvin Xu, Jimmy Ba, Ryan Kiros, Kyunghyun Cho, Aaron Courville, Ruslan
  Salakhudinov, Rich Zemel, and Yoshua Bengio.
\newblock 2015.
\newblock Show, attend and tell: Neural image caption generation with visual
  attention.
\newblock In {\em Proceedings of the 32nd International Conference on Machine
  Learning}, pages 2048--2057, Boston, Massachusetts, USA.

\bibitem[\protect\citename{Yang \bgroup et al.\egroup
  }2015]{yang-yih-meek:2015:EMNLP}
Yi~Yang, Wen-tau Yih, and Christopher Meek.
\newblock 2015.
\newblock {WikiQA}: A challenge dataset for open-domain question answering.
\newblock In {\em Proceedings of the 2015 Conference on Empirical Methods in
  Natural Language Processing}, pages 2013--2018, Lisbon, Portugal.

\bibitem[\protect\citename{Yatskar \bgroup et al.\egroup
  }2016]{yatskar2016situation}
Mark Yatskar, Luke Zettlemoyer, and Ali Farhadi.
\newblock 2016.
\newblock Situation recognition: Visual semantic role labeling for image
  understanding.
\newblock In {\em Proceedings of the IEEE Conference on Computer Vision and
  Pattern Recognition (CVPR)}, pages 5534--5542, Zurich, Switzerland.

\bibitem[\protect\citename{Zaremba \bgroup et al.\egroup }2014]{Zaremba:2014}
Wojciech Zaremba, Ilya Sutskever, and Oriol Vinyals.
\newblock 2014.
\newblock Recurrent neural network regularization.
\newblock {\em CoRR}, abs/1409.2329.

\bibitem[\protect\citename{Zhu \bgroup et al.\egroup }2015]{Zhu:ea:2015}
Yukun Zhu, Ryan Kiros, Rich Zemel, Ruslan Salakhutdinov, Raquel Urtasun,
  Antonio Torralba, and Sanja Fidler.
\newblock 2015.
\newblock Aligning books and movies: Towards story-like visual explanations by
  watching movies and reading books.
\newblock In {\em The IEEE International Conference on Computer Vision (ICCV)},
  Santiago, Chile.

\end{thebibliography}
\bibliographystyle{acl2012}

\end{document}